\documentclass[10pt, a4paper]{article}
\usepackage{lrec}
\usepackage{multibib}
\usepackage{graphicx}
\usepackage{tabularx}
\usepackage{soul}
\usepackage{arydshln}

\usepackage{epstopdf}
\usepackage[latin1]{inputenc}

\usepackage{hyperref}
\usepackage{xstring}

\title{Sources of Complexity in Semantic Frame Parsing for Information Extraction\\}

\name{Gabriel Marzinotto$^{1, 2}$, Frederic Bechet$^2$, Geraldine Damnati$^1$, Alexis Nasr$^2$}

\address{(1) Orange Labs, (2)Aix Marseille Univ, CNRS, LIF \\
         (1) Lannion France , (2) Marseille France \\
         \{gabriel.marzinotto, geraldine.damnati\}@orange.com\\
         \{frederic.bechet, alexis.nasr\}@lif.univ-mrs.fr\\}

\abstract{
This paper describes a Semantic Frame parsing System based on sequence labeling methods, precisely BiLSTM models with highway connections, for performing information extraction on a corpus of French encyclopedic history texts annotated according to the Berkeley FrameNet formalism.
The approach proposed in this study relies on an integrated sequence labeling model which jointly optimizes frame identification and semantic role segmentation and identification. The purpose of this study is to analyze the task complexity, to highlight the factors that make Semantic Frame parsing a difficult task and to provide detailed evaluations of the performance on different types of frames and sentences.
\newline  }
\begin{document}

\maketitleabstract

\section{Introduction}

Deep Neural Networks (DNN) with word embeddings have been successfully used for semantic frame parsing~\cite{hermann2014semantic}. This model extends previous approaches \cite{das2014statistical} where classifiers are trained in order to assign the best possible roles for each of the candidate spans of a syntactic dependency tree.

On the other hand, \textit{recurrent neural networks} (RNN) with Long Short Memory (LSTM) cells have been applied to several semantic tagging tasks such as \textit{slot filling} ~\cite{mesnil2015using} or even frame parsing ~\cite{hakkani2016multi,tafforeau2016joint} for Spoken Language Understanding. Currently, there is an important amount of research addressed to  optimize architecture variants of the recurrent neural networks for the different semantic tasks. In SRL the current state of the art \cite{he2017deep} uses an 8 layers bidirectional LSTM with highway connections \cite{SrivastavaGS15}, that learns directly from the word embedding representations and uses no explicit syntactic information. 

More recently, \cite{yang2017} proposed a combined approach that learns a sequence tagger and a span classifier to perform semantic frame parsing making significant performance gains on the FrameNet test dataset. 

However, there is little work in analyzing the sources of error in Semantic Frame parsing tasks. This is mainly due to the size of the SemEval07 corpus that contains 720 different frames and 754 Frame Elements (FE), with a lexicon of 3,197 triggers, for only 14,950 frame examples in the training set. Hence the size of the dataset, as well as number of examples per frame tends to be too small to perform this type of analysis. For this reason the analyses done by researchers in the domain focus mainly on the performance of their model on rare frames \cite{hermann2014semantic}. In \cite{marzinotto2018semantic} a new corpus of French texts annotated following the FrameNet paradigm \cite{baker1998berkeley} \cite{fillmore2004framenet} is introduced. This new corpus has been partially annotated using a restricted number of Frames and triggers. The purpose was to obtain a larger amount of annotated occurrences per Frame with the counterpart of a smaller amount of Frames.


In this paper we focus on analyzing the factors that make a Frame hard to predict, describing which Frames are intrinsically difficult, but also which types of frame triggers are more likely to yield prediction errors and which sentences are complex to parse.

\section{Sequence labeling model}
  \label{seq:models}

\subsection{Highway bi-LSTM approach}

\begin{figure}[htbp] 
  \includegraphics[width=1\linewidth]{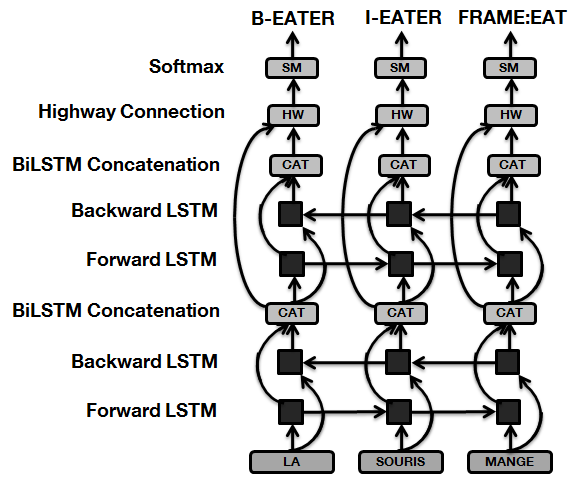}
  \caption{Highway bi-LSTM Model Diagram}
  \label{fig:hwlstm}
\end{figure}

Following the previous work of \cite{he2017deep}, in this study, we propose a similar architecture, a 4 layer bidirectional LSTM with highway connections. For this model we use two types of LSTM layers, forward ($F$) layers and backward ($B$) layers which are concatenated and propagated towards the output using highway connections \cite{SrivastavaGS15}. A diagram of our model architecture is shown in Figure \ref{fig:hwlstm}. 
There are 2 main differences between the model proposed in \cite{he2017deep}  and ours. First, we do not implement A* decoding of the output probabilities, second, our system not only relies on word embeddings as input features, but we also include embeddings encoding: syntactic dependencies, POS, morphological features, capitalization, prefixes and suffixes of the input words. We have observed these features to be useful for the FE detection and classification task.

In order to deal with both the multi-label and linking problems we have built training samples containing only one predicate. More precisely a sentence containing N predicates provides N training samples. The downside of this approach is that during prediction time, parsing a sentence with N predicates requires N model applications.
At decoding time each pair \{ sentence , predicate \} is processed by the network and a distribution probability on the frames and frame elements for each word is produced. To these probabilities we apply a \textit{coherence filter} in which we take as ground truth the frame prediction (represented as the label assigned by the tagger to the trigger) and we discard frame element labels that are incompatible to the predicted frame.

\section{The CALOR Semantic Frame Corpus}
\label{calor_corpus}

The experiments presented in this paper were carried out on the CALOR corpus, which is a compilation of documents in French that were hand annotated in frame semantics. This corpus contains documents from 4 different sources: Wikipedia's Archeology portal (WA, 201 documents), Wikipedia's World War 1 portal (WGM, 335 documents), Vikidia's~\footnote{https://fr.vikidia.org} portals of Prehistory and Antiquity (VKH, 183 documents) and ClioTexte's ~\footnote{https://clio-texte.clionautes.org/} resources about World War one (CTGM, 16 documents). 
In contrast to full text parsing corpus, the frame semantic annotations of CALOR are limited to a small subset of frames from FrameNet \cite{Baker:1998:BFP:980845.980860}. 
The goal of this \textit{partial parsing} process is to obtain, at a relatively low cost, a large corpus annotated with frames corresponding to a given applicative context. In our case this applicative context is Information Extraction (IE) from encyclopedic texts, mainly historical texts. Beyond Information Extraction, we attempt to propose new exploration paradigms through collections of documents, with the possibility to link documents not only through lexical similarity but also through similarity metrics based on \textit{semantic frame structure}. The notion of document is more central in our study than in other available corpora. This is the reason why we have chosen to annotate a larger amount of documents on a smaller amount of Frames.

Precisely, while Framenet proposes 1,223 different frames, 13,635 LUs, and 28,207 frame occurrences on full text annotations, CALOR is limited to 53 different frames, 145 LUs (among which 13 are ambiguous and can trigger at least two frames) and 21,398 frame occurrences. This means that the average number of examples per frame in CALOR is significantly higher than in the full-text annotations from FrameNet. 
\begin{table*}
\begin{center}
\scriptsize
\begin{tabular}{ccccc}
Accomplishment & Activity-start & Age & Appointing & Arrest \\
Arriving & Assistance & Attack & Awareness & Becoming \\
Becoming-aware & Buildings & Change-of-leadership & Choosing & Colonization \\
Coming-to-believe & Coming-up-with & Conduct & Contacting & Creating \\
Death & Deciding & Departing & Dimension & Education-teaching \\
Existence & Expressing-publicly & Finish-competition & Giving & Hiding-objects \\
Hostile-encounter & Hunting & Inclusion & Ingestion & Installing \\
Killing & Leadership & Locating & Losing & Making-arrangements \\
Motion & Objective-influence & Origin & Participation & Request \\
Scrutiny & Seeking & Sending & Shoot-projectiles & Statement \\
Subjective-influence & Using & Verification &  & \\
\end{tabular}
\normalsize
\end{center}
\caption{List of Semantic Frames annotated in the CALOR corpus}
\end{table*}

\section{ Results }

In order to run the experiments we divided the CALOR corpus into 80\% for training and 20\% for testing. This partition is done ensuring a similar frame distribution in training and test. 

In the CALOR corpus, ambiguity is low, with only 53 different Frames. Most triggers have only 1 possible Frame. This makes the performance of our model in the frame selection subtask as high as 97\%. For this reason we focus our analysis on the FE detection and classification subtask.

\begin{figure}[htbp] 
  \includegraphics[width=1\linewidth]{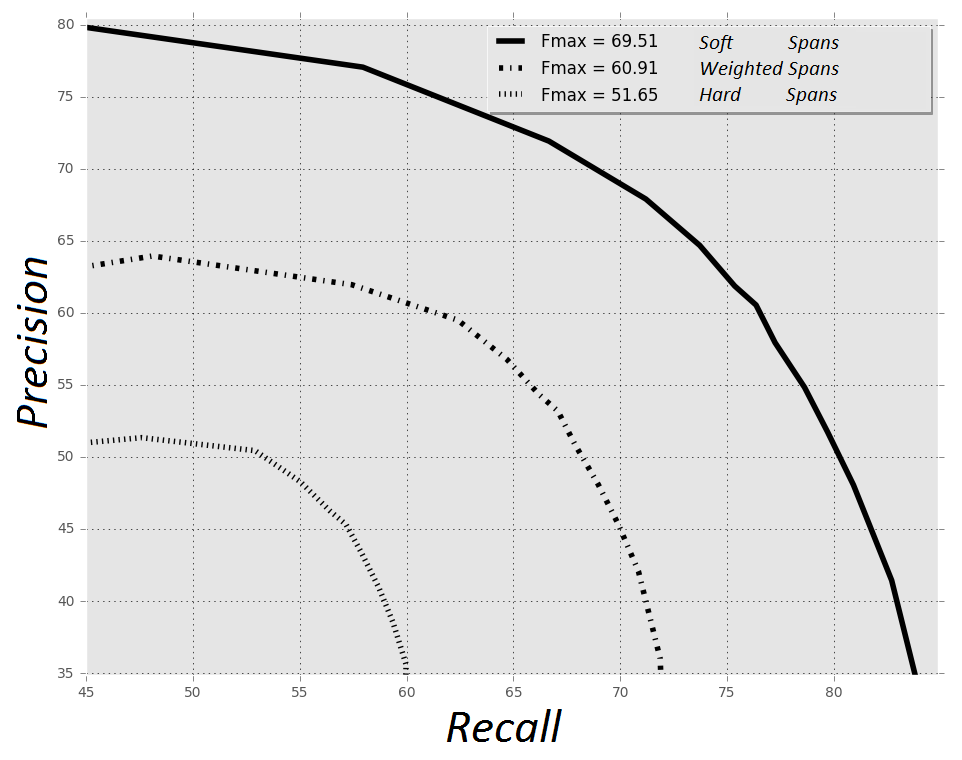}
  \caption{Model's Precision Recall curves using 3 different metrics: soft spans, weighted spans and hard spans }
  \label{fig:pr_curv}
\end{figure}

We trained our model on the CALOR corpus and we evaluated it by thresholding the output probabilities in order to build the FE detection and classification precision recall curves shown in Figure \ref{fig:pr_curv}. The three curves correspond to three possible precision-recall metrics: soft spans, weighted spans and hard spans. When evaluating using soft spans, a FE is considered correct when at least one token of its span is detected. In this case we achieve an F measure of 69,5\%. If we use the weighted span metric, a FE is scored in proportion to the size of the overlap between the hypothesis and the reference segments. Using this metric we observe a 60,9\% F-measure. Finally, the hard span metric considers a FE correct only if the full span is correctly detected. In this case, the performance degrades down to 51,7\%. This experience shows that the model detects most of the FEs but it rarely finds the full spans. It should be possible to boost the model performance by +18pts of F-measure by expanding the detected spans to its correct boundaries.

In the following subsections we present the results using the soft span metrics for the FE detection and classification task and we focus on analyzing several complexity factors in frame semantic parsing. We divide these factors into Frame Intrinsic (Section \ref{frame_intrinsic}) and Sentence Intrinsic \ref{sentence_intrinsic}). In Section \ref{document_intrinsic} we analyze the performance of the model at a document level using correlation analysis and regression techniques to retrieve relevant parameters that allow to predict the model performance on test documents.

\subsection{ Frame Intrinsic Complexity Factors }
\label{frame_intrinsic}

\begin{figure*}[htbp] 
  \centering
  \includegraphics[width=0.8\linewidth]{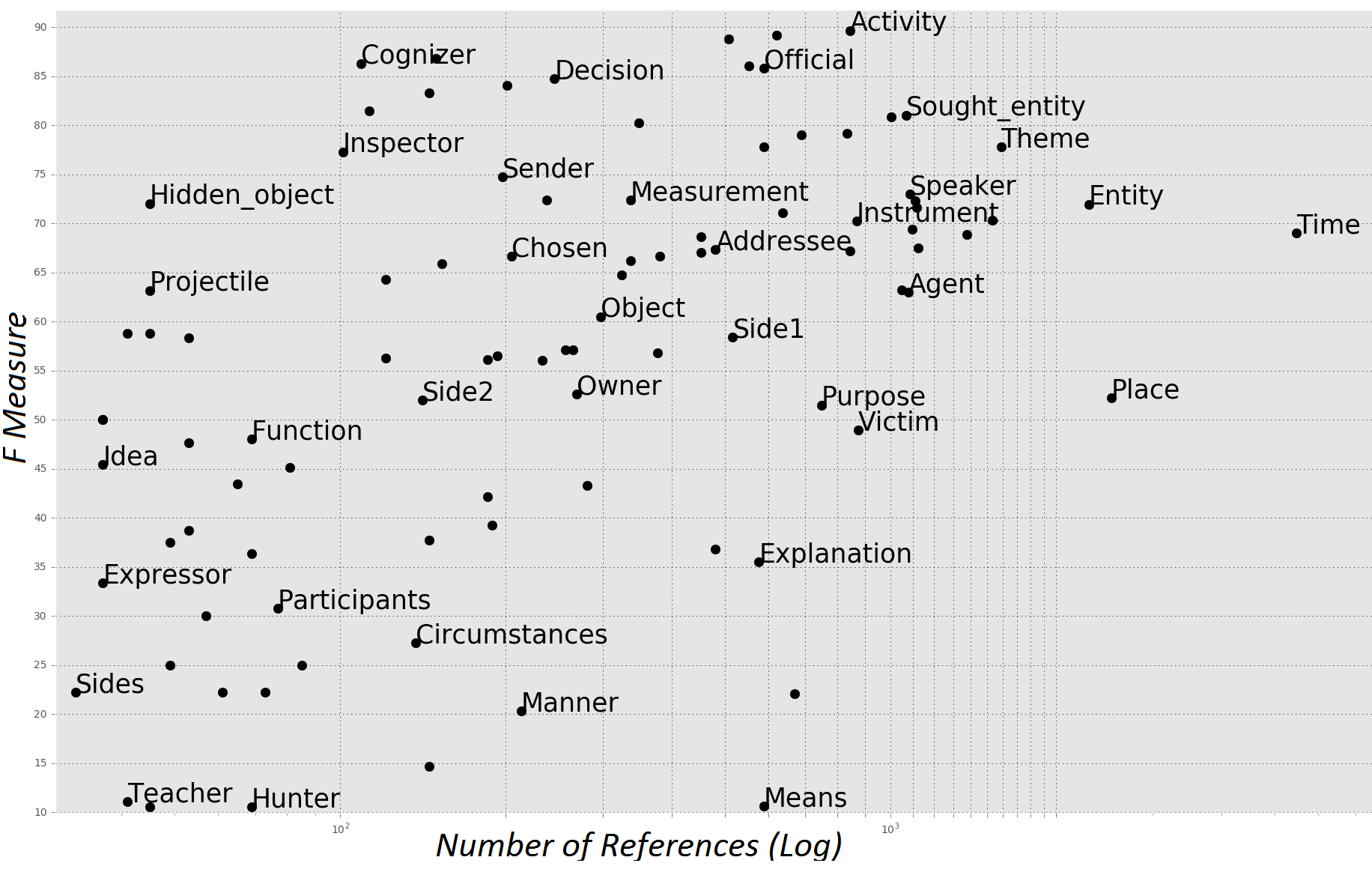}
  \caption{ Model's performance for different FEs w.r.t the number of training samples}
  \label{fig:perf_fe}
\end{figure*}

Some frames are intrinsically more difficult than others, this is due to their number of possible FEs, to the syntactic and lexical similarities between FEs and to the type of semantic concepts they represent. In Figure \ref{fig:perf_fe} we analyze the performance of our model on each FE with respect to their number of occurrences. In general, the more examples of a class we have, the better its performance should be. However, there are some ambiguity and complexity phenomena that must be taken into account.

\subsubsection{ Number of Frame Elements }

The number of possible FEs is not the same for each Frame. Intuitively, a Frame with more FEs should be harder to parse. In table \ref{table:NB_FE} we divide Frames into 3 categories \textit{Small, Medium and Large} depending of their number of possible FEs. From this experience we observed that this is not such a relevant factor and that the number of possible FEs must be really large (above 10) in order to see some degradation in the model's performance.

\begin{table}[h]
  \begin{center}
    \begin{tabular}{ c | c | c | }
      \cline{2-3}
       &  \multicolumn{1}{|p{1.8cm}|}{\centering Nb Possible FEs } & \multicolumn{1}{|p{1.8cm}|}{\centering Fmeasure} \\ \hline
      \multicolumn{1}{ |c|  }{ Small Frames}   &  1 to 7     & $ 70.5 $ \\ \hline
      \multicolumn{1}{ |c|  }{ Medium Frames}  &  8 to 10    & $ 69.8 $ \\ \hline
      \multicolumn{1}{ |c|  }{ Large Frames }  &  11 or more & $ 65.8 $ \\ \hline
    \end{tabular}
    \caption{ Performance for different Frame sizes  }
    \label{table:NB_FE}
  \end{center}
\end{table}

\subsubsection{ Specific Syntactic Realization }

Some FEs have a specific syntactic realisation. Typically, the FEs that correspond to syntactic subjects or objects (also ARG0 or ARG1 in the PropBank Paradigm). This is the case of \textit{ Activity, Official, Sought Entity, Decision, Cognizer, Inspector, Theme, Hidden Object, Expressor and Projectile }, which show good performances even when the amount of training samples is reduced. On the other hand, there are FEs such as \textit{ Time, Place, Explanation, Purpose, Manner and Circumstances }, which are realized in syntax as modifiers and have a wider range of possible instantiations. For the latter the F-measure is much lower despite of a similar amount of training samples.

\subsubsection{ Syntax Semantic Mismatch }

Some FEs syntactic realizations are different for different triggers.  
For example, the frame \textit{Education Teaching} has \texttt{\'etudier} (\textit{study}) , \texttt{enseigner}(\textit{teach}) and \texttt{apprendre} (which can translate both into \textit{learn} and \textit{teach}) as potential triggers and \textit{ Student , Teacher } among their FEs. When the trigger is \texttt{\'etudier}, the syntactic subject is \textit{Student}; when the trigger is  \texttt{enseigner}, the subject is the \textit{Teacher}; finally, when the trigger is \texttt{apprendre}, the subject could be either \textit{Student} or \textit{Teacher} and further disambiguation is needed in order to assign the correct FE. This explains the low performances observed in FEs such as \textit{Teacher}.

Some FEs are very similar up to a small nuance. For example, the Frame \textit{ Education Teaching } has 6 possible FEs to describe what is being studied \textit{ Course, Subject, Skill, Fact, Precept and Role } and their slight difference relies in the type of content studied. This kind of FE are very prone to confusions even for human annotators. Moreover, if the FE is a pronoun, finding the correct label may not be possible without the sentence context.  

Another type of FE similarity appears in symmetric actions, for example, the Frame \textit{ Hostile Encounter }, has FEs \textit{ Side1, Side2 or Sides } to describe the belligerents of an encounter. Such FEs are prone to confusions and for this reason, our model has a low performance on them.

\subsection{ Sentence Intrinsic Complexity Factors }
\label{sentence_intrinsic}

We have identified three sentence intrinsic complexity factors, the Trigger POS, the Trigger Syntactic Position and the sentence length.

\subsubsection{ Trigger POS }

As shown in table \ref{table:POS_root_Performance} the model performance varies more than 17pts of F measure depending on whether the triggers are nouns or verbs. This is due to the variety, in French, of the syntactic nature of Verb dependents when compared to Nouns. Verb arguments can be realized as Subjects, Objects, indirect Objects (introduced by specific prepositions), adverbs and Prepositional Phrases. Nouns arguments, in contrast, are usually realized as Prepositional Phrases and adjectives. Since FEs are mostly realized as arguments of their Frame trigger, Verb triggered Frames offer a wider range of syntactic, observable, means to distinguish its FE, making them easier to model. 

\subsubsection{ Trigger Syntactic Position }

We observe that the model's performance varies significantly depending on whether the sentence's triggers are at the root of the syntactic dependency tree or not. The results for this experience are summed up in Table \ref{table:POS_root_Performance}. We observe that the easiest triggers are at the root of their syntactic tree and there is a difference of 14pts of F measure between them and the triggers that occupy other positions in the syntactic tree.

\begin{table}[h]
  \begin{center}
    \begin{tabular}{ c | c | c | }
      \cline{2-3}
       &  \multicolumn{1}{|p{1.8cm}|}{\centering Percentage } & \multicolumn{1}{|p{1.8cm}|}{\centering Fmeasure} \\ \hline
      \multicolumn{1}{ |c|  }{ Verbal Trigger}    & $ 64.9 \% $ & $ 75.2 $ \\ \hdashline
      \multicolumn{1}{ |c|  }{ Nominal Trigger}   & $ 35.1 \% $ & $ 57.5 $ \\ \hline
      \multicolumn{1}{ |c|  }{ Root Trigger}      & $ 25.9 \% $ & $ 79.5 $ \\ \hdashline
      \multicolumn{1}{ |c|  }{ Non Root Trigger}  & $ 74.1 \% $ & $ 65.4 $ \\ \hline
      \multicolumn{1}{ |c|  }{ All Triggers }   & $ 100 \% $  & $ 69.5 $ \\ \hline
    \end{tabular}
    \caption{ Performance for different types of triggers  }
    \label{table:POS_root_Performance}
  \end{center}
\end{table}

\subsubsection{Sentence Length}

Another important factor is sentence length. In general, longer sentences are harder to parse as they often present more FEs and a more complex structure. Also, it is in longer sentences that we find the most non-root triggers. In this experiment our model yields 74.0\% F-measure for sentences with less than 27 words and 65.7\% F-measure for sentences with 27 words or more.
However, we also observed that parsing a long sentence (with 27 words or more) with the trigger at the root of the syntactic tree can still be done with a fairly good performance of 76.6\%. While, when parsing a long sentence with non root triggers the performances degrade down to 62.8\%. Table \ref{table:len_depth_Performance} presents the model performance with respect to trigger syntactic position and sentence length.

\begin{table}[h]
  \begin{center}
    \begin{tabular}{ c | c | c | }
      \cline{2-3}
       &  \multicolumn{1}{|p{1.8cm}|}{\centering Root Triggers } & \multicolumn{1}{|p{1.8cm}|}{\centering Non-Root Triggers} \\ \hline
      \multicolumn{1}{ |c|  }{Short ($<$ 27 words)}    & $ 81.6  $ & $ 68.2 $ \\ \hline
      \multicolumn{1}{ |c|  }{Long  ($\geq$ 27 words)} & $ 76.5  $ & $ 62.8 $ \\ \hline
    \end{tabular}
    \caption{ F-measure for different sentence lengths (Above and Below the Median) and trigger positions (Root and Non-Root) }
    \label{table:len_depth_Performance}
  \end{center}
\end{table}

All of these factors add up. Short sentences with a verbal trigger at the root of the syntactic tree can be parsed with an F-measure of 82.6\%  while long sentences with noun triggers that are not root of the syntactic tree are parsed with an F-measure of 52.8\%.

\subsection{ Document Intrinsic Complexity Factors}
\label{document_intrinsic}

In the previous sections we have presented the main factors that influence semantic frame parsing. In order to quantify the impact of each factor we have compared so far the performance of our model in subsets of the test set corresponding to the different modalities of the complexity factors. In this section, we try to directly evaluate the impact of these factors on the model performance. The analysis is performed here at the document level, the applicative motivation being to be able to predict the semantic parsing performances for a given new document. 

We address the analysis of the complexity factors as a regression problem where we describe a dependent variable \textit{y} (that quantifies the  model performance) using a set of explanatory variables \textit{ $X=(X_1,...,X_n)$} which are our candidate complexity factors.

First, we use our model to generate hypothesis predictions of frame semantic parsing on the entire CALOR corpus, with a 5-fold protocole. For each fold, we train on 80\% of the corpus, and generate predictions for the remaining 20\%.

Then, we evaluate the model's predictions using the gold annotations to compute the model's performance for each of the 735 documents in the CALOR corpus (Section \ref{calor_corpus}). For each of these documents, we also compute the set of features (complexity factors candidates) listed below:

\begin{itemize}
\setlength\itemsep{-1pt}
\item Percentage of root / non-root triggers. 
\item Percentage of verbal / nominal triggers.
\item Mean phrase length.
\item Mean trigger syntactic depth.
\item Mean trigger position in sentence. 
\item Part of Speech (POS) distribution\footnote[1]{ POS and DEP from the Universal Dependencies project (http://universaldependencies.org/)}.
\item Syntactic dependency relation (DEP) distribution\footnotemark[1].
\end{itemize}

To avoid taking into consideration parts of the document that were not processed by our semantic frame parser, these features are computed only using the sentences that contain at least 1 trigger. In order to make the analysis more robust to outliers we discarded the documents with less than 30 triggers, yielding a total amount of 327 documents. Unidimensional statistics show that the model's F-measure follows a Gaussian distribution across documents. This Gaussian distribution is centered at 69 pts of F-measure and has an standard deviation of 6.5 pts. This value of standard deviation shows that the model is fairly robust and has a stable performance across documents. 

Finally, we used this set of document's features and model's performances in two experiments:
\begin{itemize}
\setlength\itemsep{-1pt}
\item To compute the Pearson correlation coefficient between the F-measure and each feature.
\item To train a linear regression model that attempts to predict the model's performance on a document given a small set of parameters. 
\end{itemize}

\subsubsection{ Pearson Correlation }

We computed the Pearson correlation between each feature and the F-measure of the system and  verified that the correlation coefficient passes the Student's t-Test. Table \ref{table:correlation} shows the 15 parameters that have the highest absolute correlation with the F-measure. 

\begin{table}[h]
  \begin{center}
    \begin{tabular}{ c | c | c | }
      \cline{2-3}
       &  \multicolumn{1}{|p{0.8cm}|}{\centering Rank } & \multicolumn{1}{|p{1.8cm}|}{\centering Pearson Correlation} \\ \hline
      \multicolumn{1}{ |c| }{ Mean Trigger Depth }       & $ 1 $  & $ -0.44 $ \\ \hline
      \multicolumn{1}{ |c| }{ Mean Trigger Position }    & $ 2 $  & $ -0.36 $ \\ \hline
      \multicolumn{1}{ |c| }{ Verbal Trigger Percentage} & $ 3 $  & $ +0.31 $ \\ \hline
      \multicolumn{1}{ |c| }{ Mean Sentence Length }     & $ 4 $  & $ -0.30 $ \\ \hline
      \multicolumn{1}{ |c| }{ DEP Oblique Nominal }      & $ 5 $  & $ +0.30 $ \\ \hline     
      \multicolumn{1}{ |c| }{ DEP Passive Auxiliary }    & $ 6 $  & $ +0.29 $ \\ \hline
      \multicolumn{1}{ |c| }{ POS Punctuation }          & $ 7 $  & $ -0.28 $ \\ \hline
      \multicolumn{1}{ |c| }{ POS Proper Noun }          & $ 8 $  & $ +0.27 $ \\ \hline
      \multicolumn{1}{ |c| }{ POS Adverbs }              & $ 9 $  & $ -0.20 $ \\ \hline
      \multicolumn{1}{ |c| }{ Multi Words Expressions }  & $ 10 $ & $ -0.20 $ \\ \hline
      \multicolumn{1}{ |c| }{ DEP Prepositional Case}    & $ 11 $ & $ +0.20 $ \\ \hline      
      \multicolumn{1}{ |c| }{ POS Preposition }          & $ 12 $ & $ +0.15 $ \\ \hline
      \multicolumn{1}{ |c| }{ POS Conjunction }          & $ 13 $ & $ -0.14 $ \\ \hline
      \multicolumn{1}{ |c| }{ DEP Copula }               & $ 14 $ & $ -0.14 $ \\ \hline
      \multicolumn{1}{ |c| }{ POS Number }               & $ 15 $ & $ +0.11 $ \\ \hline
      \end{tabular}
    \caption{ Pearson Correlation between the best 15 Document Features and the F-measure }
    \label{table:correlation}
  \end{center}
\end{table}

In table \ref{table:correlation} we observe that the most important parameter is the syntactic depth of the trigger. As we have previously shown, triggers at the root of the syntactic tree have the best performances (also, root triggers are often verbs). The second most correlated parameter is the position of the trigger in the sentence. Triggers that are far from the beginning of the sentence show lower performance, as they are prone to errors in syntax. Our third and fourth parameters are the percentage of verbal triggers and the average sentence length. As shown in previous experiences, verbal triggers and short sentences are, in general, easier to parse.

This study also reveals morpho-syntactic parameters that are correlated with the model's performances:  documents with a large amount of punctuation marks, adverbs, and conjunctions are more complex and harder to parse. On the other hand, documents with a large proportion of proper nouns are simpler, as proper nouns correspond to places, institutions and persons' names, which often appear as Frame Elements. The same observation can be made with Numbers, that correspond to dates and quantities. Prepositions also facilitate parsing, as they are associated to specific FEs.  

As concerns dependency parsing related features, the highest correlation is observed for Oblique Nominal (OBL) dependency. OBL dependencies attach a noun phrase functioning as a non-core argument to the syntactic head. Documents with a large proportion of oblique nominal groups are positively correlated with the F-measure. OBL arguments are often annotated as FEs \texttt{ (Time, Place, Purpose...) } and when they contain a Prepositional Case they are easy to associate to their corresponding FE. This also explains the positive correlation of the Prepositional Case. Surprisingly, documents with a large proportion of sentences in passive voice are correlated with better performances, while copula verbs degrade the results. In the CALOR corpus, some copula verbs are annotated as triggers \textit{se nommer (to be named), \^etre \'elu (to be elected), devenir (to become)) }. Thus the negative correlation may be due to low performances for these lexical units. 

Finally, Multi Words Expressions (MWE) are also associated with low performances, as the meaning of unseen MWE is harder to be inferred, misleading the semantic parser.

\subsubsection{ Performance Inference }

In this experience we trained a linear regression model with incremental feature selection using cross validation. The objective is to predict the performances of our frame semantic parser on a document given a small set of parameters. 

In incremental feature selection, we start with an empty set of selected features. At each iteration of the algorithm we test all the unselected feature candidates and we pick the feature that minimizes the cross validation mean square error (MSE) given all the previously selected features. The algorithm's stopping criterion finishes the process when the MSE no longer evolves. Unlike the previous experience where all features are evaluated independently, this experimet allows to select a smaller feature set that is not redundant.

We evaluate the usefulness of our linear regression models by comparing them with a naive constant model that always predicts the average document performance observed on the training corpus. Note that the training corpus here means the training corpus for regression estimation but not for the semantic frame parsing model estimation (each document is parsed in a k-fold protocole). Incremental Feature Selection determines that the optimal linear regression model is trained using 8 features and the insertion of more parameters does not reduce the MSE. Table \ref{table:linreg} shows the MSE for the naive model \texttt{(Mean F-measure)} and compares it with each step of our linear regression with incremental feature selection. Each row in Table \ref{table:linreg} adds a new feature to the linear regression model, up to the last row that contains the final set of 8 selected parameters.

We observe that the naive prediction algorithm has a MSE of 42.7. A linear regression model with only one feature \texttt{(Mean Trigger Depth)} reduces MSE by 16\% relative and the best linear regression model with 8 features \texttt{(Mean Trigger Depth, DEP Oblique Nominal, Verbal Trigger Percentage, DEP Passive Auxiliary, DEP Copula, DEP Fixed Multi Words, POS Punctuation, POS Proper Noun)} yields a 41\% relative MSE reduction.

\begin{figure}[htbp] 
  \centering
  \includegraphics[width=1\linewidth]{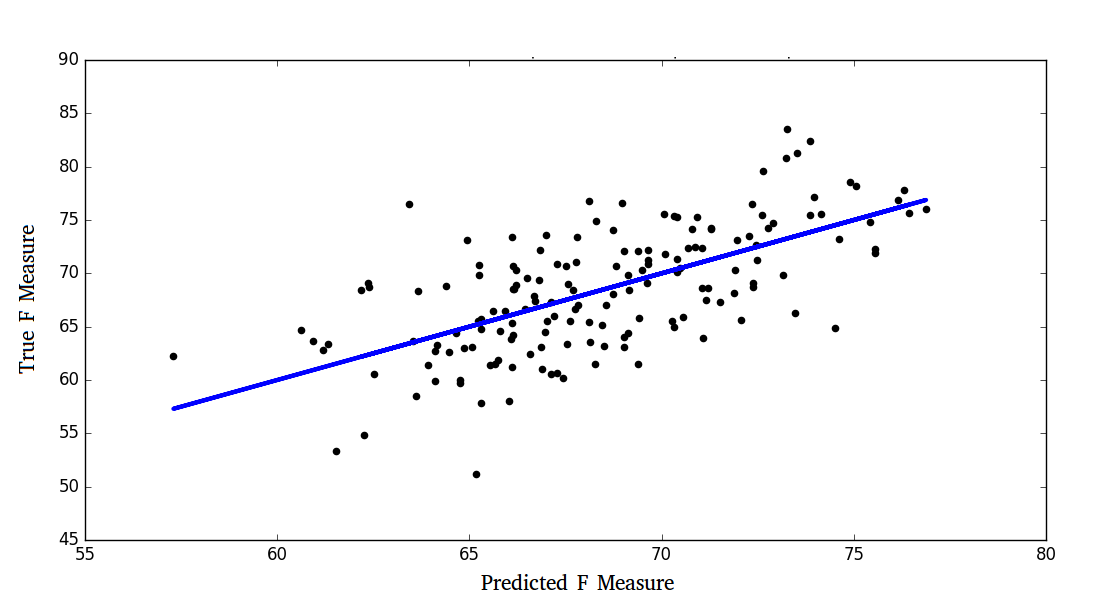}
  \caption{ Document Scatter plot showing True F-measure vs Predicted F-measure using a Linear Regression with 8 features }
  \label{fig:linreg_plot}
\end{figure}

Figure \ref{fig:linreg_plot} shows a scatter plot of the documents with their predicted F-measure and their true F-measure. We can clearly observe that both scores are correlated and the variance that can be explained by the linear regression is $ R^2=0.46$. 
However, there is still more than half of the variance that remains unexplained by the linear regression. This is because frame semantic parsing is a very complex task and the model's performances depend on many other phenomena such as the lexical coverage, lexical units and  frames that appear within a document, the type of FEs that are evocated and the degree of ambiguity at each level.

\begin{table}[h]
  \begin{center}
    \begin{tabular}{ c | c | c | }
      \cline{2-3}
       &  \multicolumn{1}{|p{1.6cm}|}{\centering \# Features } & \multicolumn{1}{|p{0.8cm}|}{\centering MSE } \\ \hline
      \multicolumn{1}{ |c| }{ Mean F-measure }           & $ 0 $  & $ 42.7 $ \\ \hline
      \multicolumn{1}{ |c| }{ Mean Trigger Depth }       & $ 1 $  & $ 35.9 $ \\ \hline
      \multicolumn{1}{ |c| }{ DEP Oblique Nominal }      & $ 2 $  & $ 33.5 $ \\ \hline     
      \multicolumn{1}{ |c| }{ Verbal Trigger Percentage} & $ 3 $  & $ 30.5 $ \\ \hline
      \multicolumn{1}{ |c| }{ DEP Passive Auxiliary }    & $ 4 $  & $ 29.1 $ \\ \hline
      \multicolumn{1}{ |c| }{ DEP Copula }               & $ 5 $  & $ 27.4 $ \\ \hline
      \multicolumn{1}{ |c| }{ Multi Words Expressions }  & $ 6 $  & $ 26.3 $ \\ \hline
      \multicolumn{1}{ |c| }{ POS Punctuation }          & $ 7 $  & $ 25.6 $ \\ \hline
      \multicolumn{1}{ |c| }{ POS Proper Noun }          & $ 8 $  & $ 25.1 $ \\ \hline
    \end{tabular}
    \caption{ Mean Squared Error (MSE) for Linear Regression with Incremental Feature Selection }
    \label{table:linreg}
  \end{center}
\end{table}

\section{Conclusion}

In this paper we proposed to identify complexity factors in Semantic Frame parsing. To do so we ran experiments on the CALOR corpus using a frame parsing model that considers the task as a sequence labeling task.
In our case only \textit{partial} annotation is considered. Only a small subset of the FrameNet lexicon is used, however the amount of data annotated for each frame is larger than in any other corpora, allowing to make more detailed evaluations of the error sources on the FE detection and classification task.
The main contribution of this work is to characterize the principal sources of error in semantic frame parsing. We divide these sources of error into two main categories: Frame intrinsic and sentence intrinsic. Examples of Frame intrinsic factors are the number of possible FE, and the syntactical similarity between them. As for the sentence intrinsic factors, we enhanced the position of the trigger in the syntactic tree, the POS of the trigger and the sentence length. In this work we showed that some morpho-syntactic categories and syntactic relations have an impact on the complexity of the frame semantic parsing. Finally, we showed that it is possible to make a fair prediction of the model's performance on a given document thanks to a regression estimation knowing its sentence intrinsic parameters. The features selected for the regression estimation confirm the observations regarding the task complexity but also enhance new assertions.
The complexity factors presented in this article may allow further work on feature engineering to improve the frame semantic parsing models.

\section{Bibliographical References}
\label{main:ref}

\bibliographystyle{lrec}
\bibliography{xample}

\end{document}